\documentclass[preprint,12pt]{elsarticle}

\usepackage{amssymb}
\usepackage{booktabs}
\usepackage{amsmath}
\usepackage{booktabs}
\usepackage{float}

\journal{arxiv}

\begin{document}

\begin{frontmatter}



\title{Zero-shot Load Forecasting for Integrated Energy Systems: A Large Language Model-based Framework with Multi-task Learning}


\author[label1]{Jiaheng Li}
\author[label1]{Donghe Li}
\author[label2]{Ye Yang}
\author[label3]{Huan Xi}
\author[label4]{Yu Xiao}
\author[label5]{Li Sun}
\author[label1]{Dou An}
\author[label1,label6]{Qingyu Yang}

\affiliation[label1]{organization={School of Automation Science and Engineering, Xi’an Jiaotong University}, addressline={}, city={Shaanxi}, postcode={710049}, country={China}}
\affiliation[label2]{organization={School of Electronic Information Engineering, Shanghai Dianji University}, addressline={300 Shuihua Road, Pudong New District}, city={Shanghai}, postcode={201306}, country={China}}
\affiliation[label3]{organization={Key Laboratory of Thermo-Fluid Science and Engineering of Ministry of Education, School of Energy and Power Engineering, Xi'an Jiaotong University}, addressline={}, city={Shaanxi}, postcode={710049}, country={China}}
\affiliation[label4]{organization={Postdoc in Electrical Engineering, Eindhoven University of Technology}, addressline={}, city={Eindhoven}, postcode={5600 MB}, country={ the Netherlands}}
\affiliation[label5]{organization={School of Microelectronics, Xi’an Jiaotong University}, addressline={}, city={Shaanxi}, postcode={710049}, country={China}}
\affiliation[label6]{organization={State Key Laboratory for Manufacturing System Engineering Lab, Xi’an Jiaotong University}, addressline={}, city={Shaanxi}, postcode={710049}, country={China}}

\begin{abstract}
The growing penetration of renewable energy sources in power systems has increased the complexity and uncertainty of load forecasting, especially for integrated energy systems with multiple energy carriers. Traditional forecasting methods heavily rely on historical data and exhibit limited transferability across different scenarios, posing significant challenges for emerging applications in smart grids and energy internet. This paper proposes the TSLLM-Load Forecasting Mechanism, a novel zero-shot load forecasting framework based on large language models (LLMs) to address these challenges. The framework consists of three key components: a data preprocessing module that handles multi-source energy load data, a time series prompt generation module that bridges the semantic gap between energy data and LLMs through multi-task learning and similarity alignment, and a prediction module that leverages pre-trained LLMs for accurate forecasting. The framework's effectiveness was validated on a real-world dataset comprising load profiles from 20 Australian solar-powered households, demonstrating superior performance in both conventional and zero-shot scenarios. In conventional testing, our method achieved a Mean Squared Error (MSE) of 0.4163 and a Mean Absolute Error (MAE) of 0.3760, outperforming existing approaches by at least 8\%. In zero-shot prediction experiments across 19 households, the framework maintained consistent accuracy with a total MSE of 11.2712 and MAE of 7.6709, showing at least 12\% improvement over current methods. The results validate the framework's potential for accurate and transferable load forecasting in integrated energy systems, particularly beneficial for renewable energy integration and smart grid applications.
\end{abstract}




\end{frontmatter}



\section{Introduction}
The growing penetration of renewable energy generation has led to significant challenges for power systems, particularly in terms of system dispatch and balance. The inherent variability, intermittency, and unpredictability of renewable energy sources make traditional power system management techniques increasingly inadequate. In the development of integrated energy systems that incorporate multiple energy forms, the effective management of multi-energy flexible loads—such as electricity, heating/cooling, and natural gas at the consumer end—has emerged as a key strategy for mitigating renewable energy instability at the production end\cite{1,2,3,4}.

\begin{figure}[!t]
\centering
\includegraphics[width=\columnwidth]{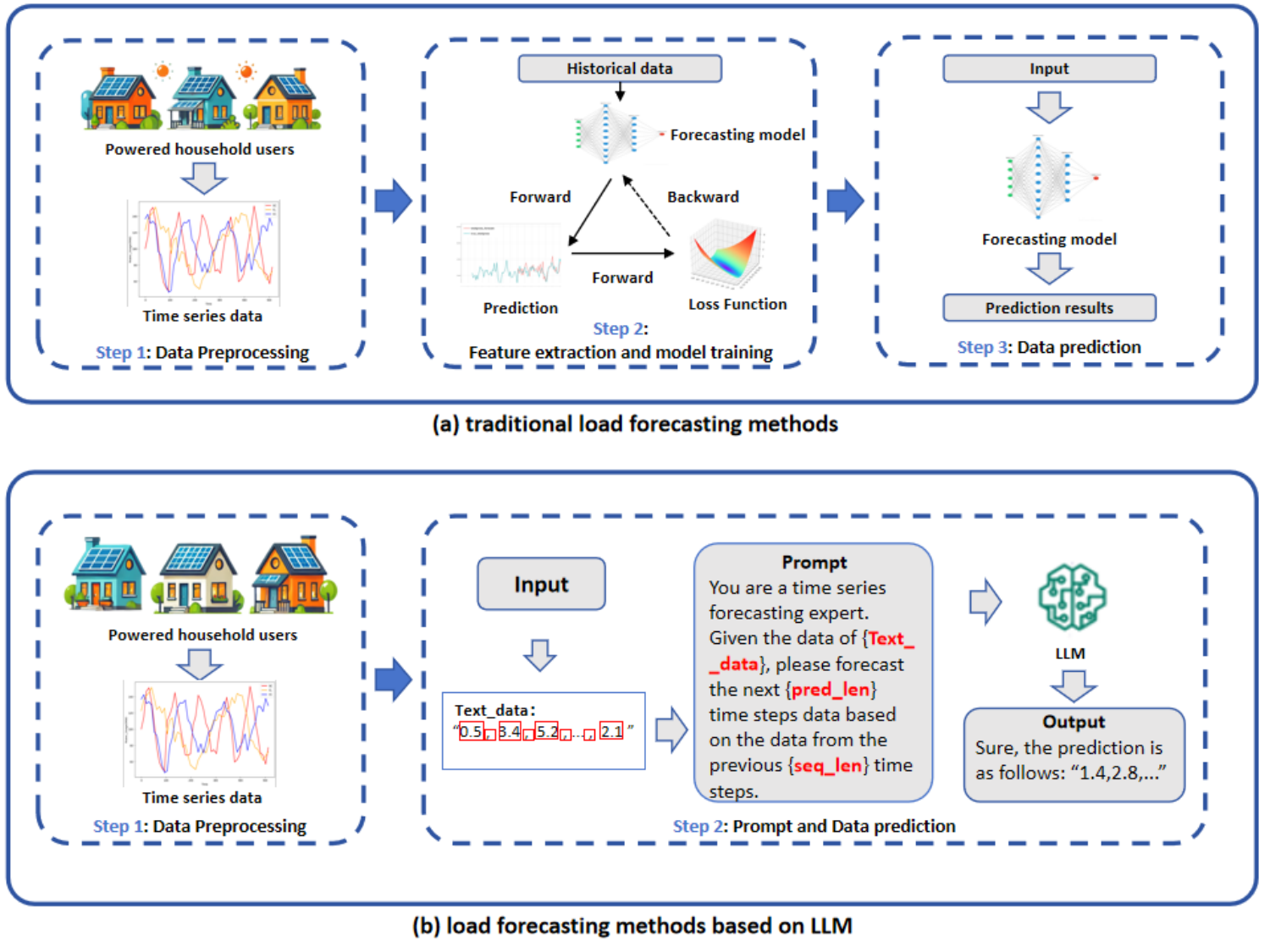}
\caption{Comparison of load forecasting schemes: (a) Traditional machine learning methods and (b) Large language model-based approach. The traditional method requires extensive historical data and exhibits limited transferability, while the LLM-based method enables zero-shot prediction across different data features.}
\label{fig1}
\end{figure}

Load forecasting plays a crucial role in energy management systems, facilitating power system planning and operational management. It helps predict future electricity demand, ensuring a balance between power supply and demand. This balance is essential for the efficient distribution and utilization of energy within the energy internet framework. As global energy structures evolve and integrated energy system dispatching advances, load forecasting technologies have gained significant attention. These technologies have reached a mature stage, encompassing various forecasting models and techniques, such as regression models, time series models\citep{5,6,7}, deep learning models\citep{8,9,10,11,12,13,14}, and artificial intelligence (AI) models\citep{15,16,17,18,19}. Accurate load forecasting is critical for ensuring grid stability, optimizing resource allocation, promoting energy complementarity, and advancing the construction of the energy internet and integrated energy systems. Additionally, it plays a pivotal role in realizing the development of the Internet of Things (IoT) across the power industry\cite{20}.

With the rapid advancements in artificial intelligence, time series forecasting—a major branch of AI—has been widely applied in load forecasting. For example, several studies have explored the use of convolutional neural networks (CNN) and long short-term memory (LSTM) networks for short-term load forecasting in cogeneration power systems\cite{21,22}. Temporal Convolutional Networks (TCN) combined with LSTM models have also been successfully applied for extracting temporal features from raw load data, showing improved forecasting accuracy\cite{23}. Despite the success of these traditional methods, deep learning models typically require large amounts of historical data for feature extraction and model training, which limits their effectiveness in cold-start scenarios where limited historical data is available.

Given the data dependency of traditional deep learning algorithms in load forecasting, research has increasingly focused on developing forecasting models that perform well under few-shot or zero-shot conditions. The emergence of large language models (LLMs) has provided a promising solution to this challenge. LLMs, such as GPT-2 with 1.5 billion parameters, are deep learning models trained on vast amounts of text data and capable of performing a wide range of natural language processing (NLP) tasks, including text generation, translation, question answering, and sentiment analysis\cite{24,25}. Their strong performance across general tasks has led to their exploration in time series forecasting tasks\cite{26,27,28}. For example, experiments have shown that pre-trained LLMs can function effectively as zero-shot time series forecasters\cite{29}. Beyond simply using LLMs as tools for time series forecasting\cite{30,31}, researchers have focused on methods that enable LLMs to understand time series data. These include techniques for encoding time series data\cite{32} and aligning it within a semantic space\cite{33,34,35,36,37,38}. Further studies have employed multimodal large models for time series forecasting, including methods that transform time windows into images and use visual encoders to map these images into a form understandable by LLMs\cite{39}. Additionally, other research efforts have aimed at translating time series data into a text format suitable for LLMs, which has significantly improved the accuracy of forecasting models\cite{40}.

In this context, this paper proposes the TSLLM-Load Forecasting Mechanism, a novel zero-shot load forecasting framework based on LLMs, aimed at addressing the challenges of transferability and the complexities associated with understanding time series data. Specifically, this approach tackles three key issues: the variability in household characteristics, the transformation of load data into a textual format that can be processed by LLMs, and the interdependencies among multiple features within the data.

The contributions of this paper are as follows:

\begin{itemize}
\item \textbf{A Transferable Zero-shot Time Series Forecasting Framework Based on LLMs:} We introduce the TSLLM-Load Forecasting Mechanism, a comprehensive framework for household electricity load forecasting centered on LLMs. This framework includes a data preprocessing module, a time series prompt generation module, and a forecasting module based on pre-trained LLMs. The data preprocessing module processes the data, while the time series prompt generation module converts the data into a format understandable by LLMs. The pre-trained LLM is then used to generate accurate forecasts, which are reconstructed into time series data through the model's output layer.

\item \textbf{Optimization of Time Series Forecasting for LLMs:} In the preprocessing module, we clean, interpolate, and normalize the data from integrated energy households. The time series prompt generation module employs a multi-task learning mechanism to capture correlations between various features of the time series data. The data undergoes trend-seasonal decomposition (SFT) before being segmented to accommodate the LLM’s context length limitations. Each feature’s data block is projected into a text domain understood by the LLM, and similarity matching is employed for text alignment. The aligned text serves as the prefix prompt for the time series data, which is then input into the LLM for forecasting. During the pre-training phase, a small subset of data is used to train certain parameters of the LLM. The full set of electricity household data is used to validate the model’s performance.

\item \textbf{Extensive Experimental Validation:} We evaluate the performance of the TSLLM-Load Forecasting Mechanism using real-world data, including a dataset of 20 Australian solar-powered households' electricity load data. In conventional testing, our method achieved a Mean Squared Error (MSE) of 0.4163 and a Mean Absolute Error (MAE) of 0.3760, outperforming other popular methods by at least 8\%. In zero-shot prediction experiments, the framework maintained consistent accuracy across 19 households, with a total MSE of 11.2712 and MAE of 7.6709, representing at least a 12\% improvement over existing methods. These results demonstrate the effectiveness of the proposed framework for accurate and transferable load forecasting, particularly in the context of renewable energy integration and smart grid applications.
\end{itemize}

The remainder of this paper is organized as follows: Section 2 presents the system model and problem formulation. Section 3 introduces the TSLLM-Load Forecasting Mechanism and its components. Section 4 discusses the experimental results, and Section 5 concludes the paper.

\section{System Model of Load Forecasting}

This section establishes a comprehensive mathematical framework for energy load forecasting in integrated energy systems. We first formalize the load forecasting problem for systems with multiple energy sources and then outline the key challenges in developing effective prediction models.

In modern integrated energy systems, multiple energy sources contribute to meeting diverse user demands. The Energy Management System (EMS) is responsible for forecasting energy requirements to optimize resource utilization, minimize waste, and guide rational electricity consumption. Consider a system with $N$ users, denoted as $\mathcal{U} = \{u_1, u_2, ..., u_N\}$. For each user $u_i$, their energy consumption data over a specific time period can be represented by a multi-dimensional time series $\mathbf{X}_i^f$, where $f$ indicates different energy features, and $t$ represents the temporal dimension. For a given time interval $\Delta t$, the data for user $u_i$ can be expressed as:

\begin{equation}
\mathbf{X}_i(\Delta t) = \{\mathbf{X}_i^{hp}, \mathbf{X}_i^{ap}, \mathbf{X}_i^{sp}\} \in \mathbb{R}^{T \times 3}
\end{equation}

where:
\begin{itemize}
\item $\mathbf{X}_i^{hp} \in \mathbb{R}^T$: Power consumption of high-power appliances
\item $\mathbf{X}_i^{ap} \in \mathbb{R}^T$: Power consumption of other appliances
\item $\mathbf{X}_i^{sp} \in \mathbb{R}^T$: Power generation from solar installations
\end{itemize}

The net energy consumption can be formulated as a temporal sequence:

\begin{equation}
\mathbf{E}_i(\Delta t) = \mathbf{X}_i^{hp} + \mathbf{X}_i^{ap} - \mathbf{X}_i^{sp}
\end{equation}

For a time interval $\Delta t$ containing $T$ time steps, the temporal evolution of energy consumption is represented as:

\begin{equation}
\mathbf{E}_i(\Delta t) = \{\mathbf{e}_i^1, \mathbf{e}_i^2, ..., \mathbf{e}_i^T\} \in \mathbb{R}^T
\end{equation}

The complete energy consumption profile for all users during this period forms a multi-dimensional tensor:

\begin{equation}
\mathbf{E}(\Delta t) = \{\mathbf{E}_1(\Delta t), \mathbf{E}_2(\Delta t), ..., \mathbf{E}_N(\Delta t)\} \in \mathbb{R}^{N \times T}
\end{equation}

\begin{equation}
\mathbf{E}(\Delta t) = \{\mathbf{e}_i^t | i \in [1,N], t \in [1,T]\}
\end{equation}

The objective of load forecasting is to predict energy consumption $\mathbf{\hat{E}}(\Delta t_{+k})$ for the next $k$ time intervals:

\begin{equation}
\mathbf{\hat{E}}(\Delta t_{+k}) = f(\mathbf{E}(\Delta t)) + \boldsymbol{\epsilon}
\end{equation}

where $f(\cdot)$ represents the forecasting function and $\boldsymbol{\epsilon}$ denotes the prediction error term. For individual users, this forecasting task can be expressed as:

\begin{equation}
\mathbf{\hat{E}}_i(\Delta t_{+k}) = f_i(\mathbf{E}_i(\Delta t)) + \boldsymbol{\epsilon}_i
\end{equation}

This formulation presents several significant challenges:

\begin{itemize}
\item \textbf{Stochastic Nature}: Both electrical appliance usage and solar power generation exhibit inherent randomness, characterized by time-varying probability distributions $P(X_t|X_{t-1}, ..., X_1)$.

\item \textbf{Heterogeneous Patterns}: Energy consumption behaviors vary significantly across users due to local conditions, installed equipment, and individual habits, resulting in distinct data distributions $P_i(X) \neq P_j(X)$ for $i \neq j$.

\item \textbf{Temporal Dynamics}: Usage patterns and solar generation exhibit strong temporal dependencies, while maintaining non-stationarity across different time scales, expressed as $\mathbb{E}[X_t] \neq \mathbb{E}[X_{t+\tau}]$ for some time lag $\tau$.
\end{itemize}

These challenges necessitate a robust and adaptable forecasting framework that can effectively handle diverse data patterns while maintaining prediction accuracy across different scenarios.

\section{TSLLM-Load Forecasting Mechanism}

This section presents the proposed TSLLM-Load Forecasting Mechanism, encompassing a comprehensive framework that integrates data preprocessing, prompt generation, and prediction capabilities. The following subsections detail the design rationale, operational workflow, and mathematical foundations of each functional module.

\subsection{Design Rationale}

Building upon the load prediction analysis presented in Section 2, the implementation of Large Language Models (LLMs) for solar household load prediction presents three fundamental technical challenges beyond the inherent stochasticity common to all prediction methods. 

The primary challenge stems from the heterogeneity in household characteristics, as solar households exhibit diverse electricity consumption patterns influenced by geographical locations, climatic conditions, and solar panel installations. Traditional data-driven methodologies, which heavily rely on historical data for feature extraction, demonstrate limited efficacy in cold-start scenarios and cross-user transfers under few-shot conditions. Developing a model capable of accurate predictions across diverse solar households under few-shot conditions remains a fundamental challenge.

The second challenge involves the dimensional transformation between time series data and the textual domain. The system input comprises time series electricity load data for individual households over specified periods, denoted as $\mathbf{X}_{t} \in \mathbb{R}^{n \times d}$, where $n$ represents the sequence length of the time series (i.e., the number of time steps), and $d$ denotes the dimensionality of features at each time step. However, LLMs require textual inputs, processing them through Transformer architectures that encode relationships between text segments as high-dimensional vectors. Given that different pre-trained LLMs accommodate varying vector dimensions - with BERT and GPT-2 accepting 768-dimensional vectors and advanced models like LLAMA3 handling 1024-dimensional vectors - we propose a mapping function $f: \mathbb{R}^{n \times d} \rightarrow \mathbb{R}^{d'}$, where $d'$ represents the target embedding dimension of the specific LLM, that generates LLM-compatible input $\mathbf{E}_{t}$.

The third challenge arises from the inherent interdependencies in multivariate time series data. For a system input containing multiple interrelated features, we define:

\begin{equation}
\mathbf{X}_{t} = [\mathbf{x}^{(1)}_{t}, \mathbf{x}^{(2)}_{t}, \mathbf{x}^{(3)}_{t}]
\end{equation}

where $\mathbf{x}^{(1)}_{t} \in \mathbb{R}^n$ represents the solar power generation sequence, $\mathbf{x}^{(2)}_{t} \in \mathbb{R}^n$ denotes the major appliance power consumption sequence, and $\mathbf{x}^{(3)}_{t} \in \mathbb{R}^n$ indicates the regular appliance power consumption sequence. These features exhibit complex interdependencies characterized by the functional relationship:

\begin{equation}
f(\mathbf{x}^{(1)}_{t}, \mathbf{x}^{(2)}_{t}, \mathbf{x}^{(3)}_{t}) \mapsto \mathbb{R}
\end{equation}

where the mapping $f$ captures the intricate interactions between different types of power consumption and generation patterns across time.

\subsection{Overview}

\begin{figure}[!t]
\centering
\includegraphics[width=\linewidth]{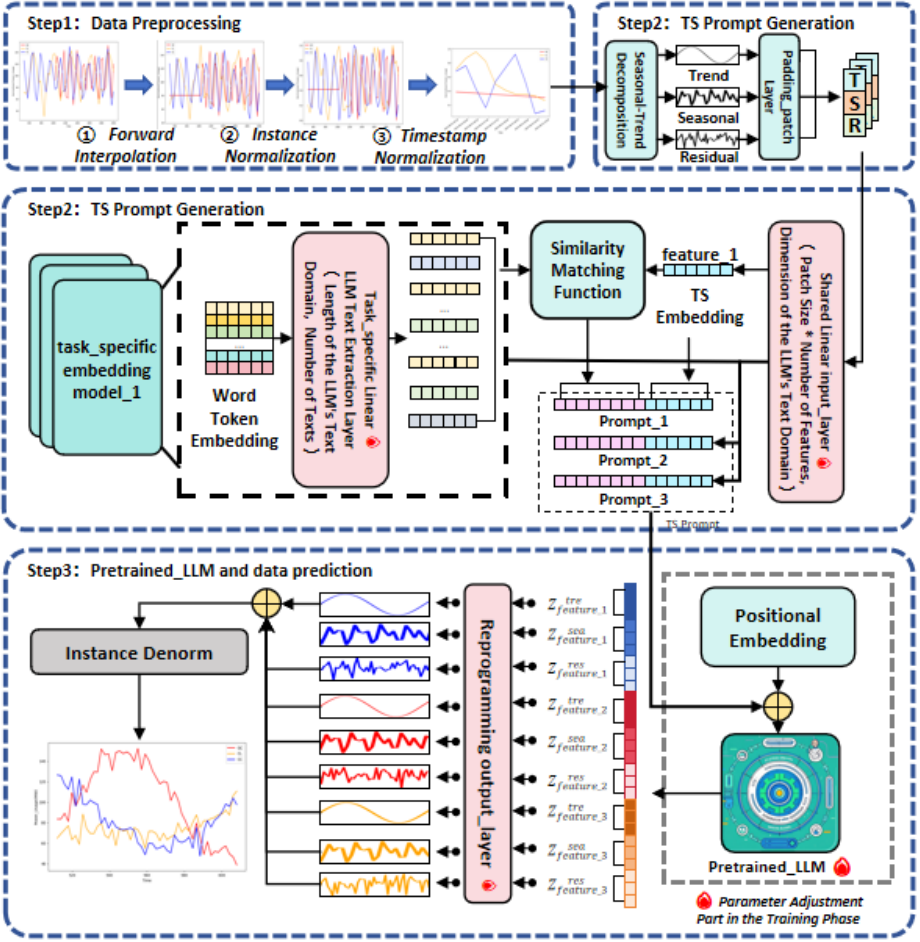}
\caption{The workflow of TSLLM-Load Forecasting Mechanism, illustrating the systematic integration of data preprocessing, prompt generation, and prediction modules.}
\label{fig:system_workflow}
\end{figure}

As illustrated in Fig.~\ref{fig:system_workflow}, the TSLLM-Load Forecasting Mechanism implements a systematic workflow through three primary functional modules. Each module performs specific transformations in the sequential process of converting raw load data into accurate forecasting results.

In Step 1, the Data Preprocessing Module establishes the foundation of our framework. This fundamental module transforms input load data into usable, normalized time series data through three key operations including forward interpolation to address missing data points while maintaining data continuity, normalization procedures to standardize the data distribution, and timestamp standardization to align temporal sequences.

In Step 2, the Time Series Prompt Generation Module serves as the critical bridge between preprocessed data and forecasting capabilities. This module processes the normalized time series load data through two sophisticated submodules. The Data Decomposition and Segmentation Module applies Seasonal and Trend decomposition (SFT) to separate normalized time series load data into trend ($\mathbf{T}_t \in \mathbb{R}^n$), periodic ($\mathbf{S}_t \in \mathbb{R}^n$), and noise ($\mathbf{R}_t \in \mathbb{R}^n$) components, and subsequently segments this decomposed data into structured blocks of size $b$ to address LLM context length limitations.

The Time Series Feature Embedding Module further processes this segmented data through three primary components:
\begin{itemize}
\item Shared Linear Mapping Layer: Implements transformation with dimensions $(b \times d, d')$ to project time series data into the LLM's textual domain
\item Task-Independent Text Extraction Layer: Operates as a linear mapping layer to extract representative tokens from the LLM's textual domain for different features
\item Alignment Module: Processes both TS-Embedding and extracted tokens through similarity function computations to construct the final prediction prompt
\end{itemize}

In Step 3, the Time Series LLM Prediction Module culminates the process by utilizing the time series prediction prompt as input and generating forecasts. This concluding module integrates three essential components:
\begin{itemize}
\item Pre-trained Large Language Model: Processes time series prediction prompts to generate forecasts for specified intervals
\item Reconstruction Output Layer: Transforms text-based predictions into structured time series data $\hat{\mathbf{X}}_{t+1} \in \mathbb{R}^{n \times d}$
\item Training and Optimization Component: Adjusts system parameters through carefully designed loss functions to achieve optimal prediction accuracy
\end{itemize}

Through this systematic integration of preprocessing, prompt generation, and prediction capabilities, the TSLLM-Load Forecasting Mechanism establishes a robust framework for zero-shot load forecasting. The structured workflow ensures consistent performance while maintaining adaptability to diverse load patterns and prediction scenarios. In the subsequent sections, we provide a detailed examination of each module's mathematical foundations, algorithmic implementations, and operational principles, beginning with the data preprocessing techniques in Section 3.3, followed by an in-depth analysis of the prompt generation mechanism in Section 3.4, and concluding with a comprehensive discussion of the prediction module in Section 3.5.

\subsection{Data Preprocessing Implementation}

The TSLLM-Load Forecasting Mechanism utilizes comprehensive load data from 20 selected solar households within the Ausgrid supply area, spanning from July 1, 2010, to June 30, 2013. The dataset encompasses households equipped with gross metering solar systems, with rigorous data quality assurance performed by the provider to exclude extreme cases of annual household electricity consumption and solar power generation efficiency during the initial year. The temporal resolution maintains consistency at 30-minute intervals, where each timestamp (e.g., 12:00 AM) represents the aggregated load data for the preceding 30-minute period (e.g., 11:30 PM to 12:00 AM).

The preprocessing pipeline comprises several essential stages to ensure data quality and consistency. For missing data points, we employ linear interpolation to maintain temporal continuity. To address the non-stationarity inherent in real-world data, we implement reversible instance normalization on the input load data. For a given load input sequence $\mathbf{X}_t$ over a time window of length $T$, the normalized sequence $\hat{\mathbf{X}}_t$ is computed through the transformation:

\begin{equation}
\hat{\mathbf{X}}_t = \gamma \cdot \frac{\mathbf{X}_t - \mu_t}{\sqrt{\sigma^2_t + \epsilon}} + \beta
\label{eq:normalization}
\end{equation}

where:
\begin{itemize}
\item $\mu_t$ and $\sigma^2_t$ represent the mean and variance computed over the temporal dimension of the specific instance respectively
\item $\gamma$ and $\beta$ are learnable parameters that enable adaptive scaling of the normalized data
\item $\epsilon$ denotes a small constant (typically $10^{-5}$) introduced to ensure numerical stability
\end{itemize}

This normalization approach ensures consistent data distribution across different time periods while preserving the relative patterns and relationships within the load sequences. The trainable parameters $\gamma$ and $\beta$ enable the model to learn optimal scaling factors during the training process, enhancing its ability to adapt to varying load patterns.

The implementation of this preprocessing module adheres to strict quality control measures. The linear interpolation for missing values considers both temporal proximity and pattern consistency to maintain data integrity. The normalization procedure is applied independently to each feature dimension, ensuring that the distinct characteristics of different load components are preserved while achieving consistent scale across the dataset.

Furthermore, the module incorporates robust error handling mechanisms and validation checks to ensure the reliability of the preprocessed data. These measures include:
\begin{itemize}
\item Boundary condition verification for interpolated values
\item Statistical consistency checks for normalized sequences
\item Temporal alignment validation for synchronized data streams
\end{itemize}

The preprocessed data provides a standardized and reliable foundation for subsequent feature extraction and model training processes. The effectiveness of these preprocessing steps is critical for the overall performance of the TSLLM-Load Forecasting Mechanism, particularly in maintaining prediction accuracy across diverse household load patterns.

\subsection{Time Series Prompt Generation Module}

The Time Series Prompt Generation Module incorporates a multi-task learning framework to effectively capture and process the intricate correlations between various features in time series data. This section presents a detailed exposition of the data decomposition, segmentation, and feature embedding methodologies.

\subsubsection{Data Decomposition and Segmentation}
We implement additive seasonal-trend decomposition to systematically partition the normalized time series into its fundamental constituent components. The decomposition process is mathematically formulated as:

\begin{equation}
\mathbf{X}_t = \mathbf{T}_t + \mathbf{S}_t + \mathbf{R}_t
\label{eq:decomposition}
\end{equation}

where $\mathbf{T}_t \in \mathbb{R}^n$, $\mathbf{S}_t \in \mathbb{R}^n$, and $\mathbf{R}_t \in \mathbb{R}^n$ represent the long-term trend, seasonal, and residual components, respectively. The decomposition follows a classical seasonal-trend additive methodology. Initially, the long-term trend component is extracted utilizing moving average methods to capture gradual variations in the time series. Subsequently, the detrended time series undergoes analysis with predefined seasonal parameters to estimate the periodic component. Finally, the residual component, which captures stochastic fluctuations, is derived by subtracting the estimated trend and seasonal components from the normalized time series.

To effectively capture the temporal encoding and local context within the input time series, we aggregate consecutive time steps into overlapping patch tokens. For the trend component sequence $\mathbf{T}_t$, the patch token representation $\mathbf{P}_t$ is generated as follows:

\begin{equation}
\mathbf{P}_t = \{\mathbf{p}_i\}_{i=1}^{n_p}, \quad \mathbf{p}_i \in \mathbb{R}^{l \times d}
\label{eq:patching}
\end{equation}

where $l$ denotes the patch length determining the temporal granularity of analysis, $n_p$ represents the total number of patches dictating the resolution of representation, and $s_h$ indicates the horizontal sliding stride controlling the overlap between adjacent patches.

\subsubsection{Time Series Feature Embedding}

\paragraph{Multi-Task Learning Framework}
The module employs a sophisticated multi-task learning approach to process solar household data characterized by three distinct features: solar power generation, major appliance power consumption, and regular appliance power consumption. For a time series dataset with these heterogeneous features, we define:

\begin{equation}
\mathbf{X}_t = [\mathbf{x}_t^{(1)}, \mathbf{x}_t^{(2)}, \mathbf{x}_t^{(3)}]
\label{eq:features}
\end{equation}

where $\mathbf{x}_t^{(1)}, \mathbf{x}_t^{(2)}, \mathbf{x}_t^{(3)} \in \mathbb{R}^n$ represent the respective feature sequences. The initial transformation through the input model and subsequent transformation of LLM outputs are conceptualized as tasks with inherent feature similarity. The embeddings of different features are processed as distinct but interconnected sub-tasks:

\begin{equation}
\mathbf{E}_t = [\mathbf{e}_t^{(1)}, \mathbf{e}_t^{(2)}, \mathbf{e}_t^{(3)}]
\label{eq:embeddings}
\end{equation}

\paragraph{Feature Projection and Alignment}
To facilitate comprehensive feature representation, we concatenate the component tokens into meta-tokens:

\begin{equation}
\mathbf{M}_t = \text{concat}[\mathbf{T}_t, \mathbf{S}_t, \mathbf{R}_t]
\label{eq:meta_tokens}
\end{equation}

These meta-tokens undergo transformation through a shared linear projection layer:

\begin{equation}
\mathbf{E}_t = \mathbf{W}_s\mathbf{M}_t + \mathbf{b}_s
\label{eq:projection}
\end{equation}

where $\mathbf{W}_s \in \mathbb{R}^{d' \times (3d)}$ and $\mathbf{b}_s \in \mathbb{R}^{d'}$ are learnable parameters that facilitate dimensionality transformation and feature extraction. The resulting embeddings are then separated by features to maintain their distinct identities while preserving inter-feature relationships.

The token embeddings space of the pre-trained LLM is denoted as $\mathbb{R}^{V \times d}$, where $V$ represents the vocabulary size (typically in the order of tens of thousands) and $d$ is the embedding dimension. To optimize computational efficiency and enhance model performance, we employ a task-independent linear mapping layer $\mathbf{W}_t \in \mathbb{R}^{k \times V}$ to derive a subset $\mathbb{R}^{k \times d}$ for alignment with time series data, where $k \ll V$ represents a significantly reduced vocabulary size.

The alignment between semantic space and time series data utilizes cosine similarity, a metric particularly effective for high-dimensional vector spaces:

\begin{equation}
s(\mathbf{e}, \mathbf{v}) = \frac{\mathbf{e}^\top\mathbf{v}}{\|\mathbf{e}\|\|\mathbf{v}\|}
\label{eq:similarity}
\end{equation}

where $\mathbf{e} \in \mathbb{R}^d$ represents time series embeddings and $\mathbf{v} \in \mathbb{R}^d$ denotes token embeddings in the reduced vocabulary space. Based on computed similarity scores, the most semantically relevant tokens are selected to enhance the input embeddings:

\begin{equation}
\hat{\mathbf{E}}_t = \text{concat}[\{\mathbf{v}_i\}_{i=1}^k, \mathbf{E}_t]
\label{eq:enhanced_embedding}
\end{equation}

where $k$ denotes the number of selected tokens and $\mathbf{v}_i$ represents the top-$k$ similar token embeddings that maximize the semantic alignment between textual and time series domains.

\subsection{Time Series LLM Prediction Module}

This module leverages the time series prediction prompt to generate forecasts through the pre-trained large language model. It integrates prediction generation, output reconstruction, and parameter optimization in a cohesive framework to achieve accurate load forecasting.

\subsubsection{Reconstruction Output Layer and Pre-trained LLM}

After processing through the pre-trained LLM, we strategically discard portions of the model's output $\mathbf{H}_t \in \mathbb{R}^{L \times d}$ from the $m$-th layer corresponding to the prefix prompt, where $L$ represents the sequence length and $d$ denotes the embedding dimension. The remaining output undergoes flattening and transformation through a linear mapping layer to produce the model's final output:

\begin{equation}
\mathbf{Y}_t = \mathbf{W}_r\mathbf{H}_t + \mathbf{b}_r
\label{eq:reconstruction}
\end{equation}

where $\mathbf{W}_r \in \mathbb{R}^{(n \times d) \times d}$ and $\mathbf{b}_r \in \mathbb{R}^{n \times d}$ are learnable parameters of the reconstruction layer that facilitate the transformation from the LLM's embedding space to the target time series space. Owing to the initial decomposition step, the overall prediction combines individual component predictions in an additive manner. The output $\mathbf{Y}_t$ is further decomposed into its constituent components:

\begin{equation}
\mathbf{Y}_t = [\mathbf{y}_t^{(T)}, \mathbf{y}_t^{(S)}, \mathbf{y}_t^{(R)}]
\label{eq:output_decomposition}
\end{equation}

where $\mathbf{y}_t^{(T)}$, $\mathbf{y}_t^{(S)}$, and $\mathbf{y}_t^{(R)}$ represent the predicted trend, seasonal, and residual components, respectively. The final prediction is obtained through component recomposition following the inverse operation of the initial decomposition:

\begin{equation}
\hat{\mathbf{X}}_{t+1} = \mathbf{y}_t^{(T)} + \mathbf{y}_t^{(S)} + \mathbf{y}_t^{(R)}
\label{eq:final_prediction}
\end{equation}

This additive recombination preserves the distinct characteristics of each component while yielding a comprehensive forecasting result.

\subsubsection{Training and Optimization Strategy}

\begin{figure}[!t]
\centering
\includegraphics[width=\linewidth]{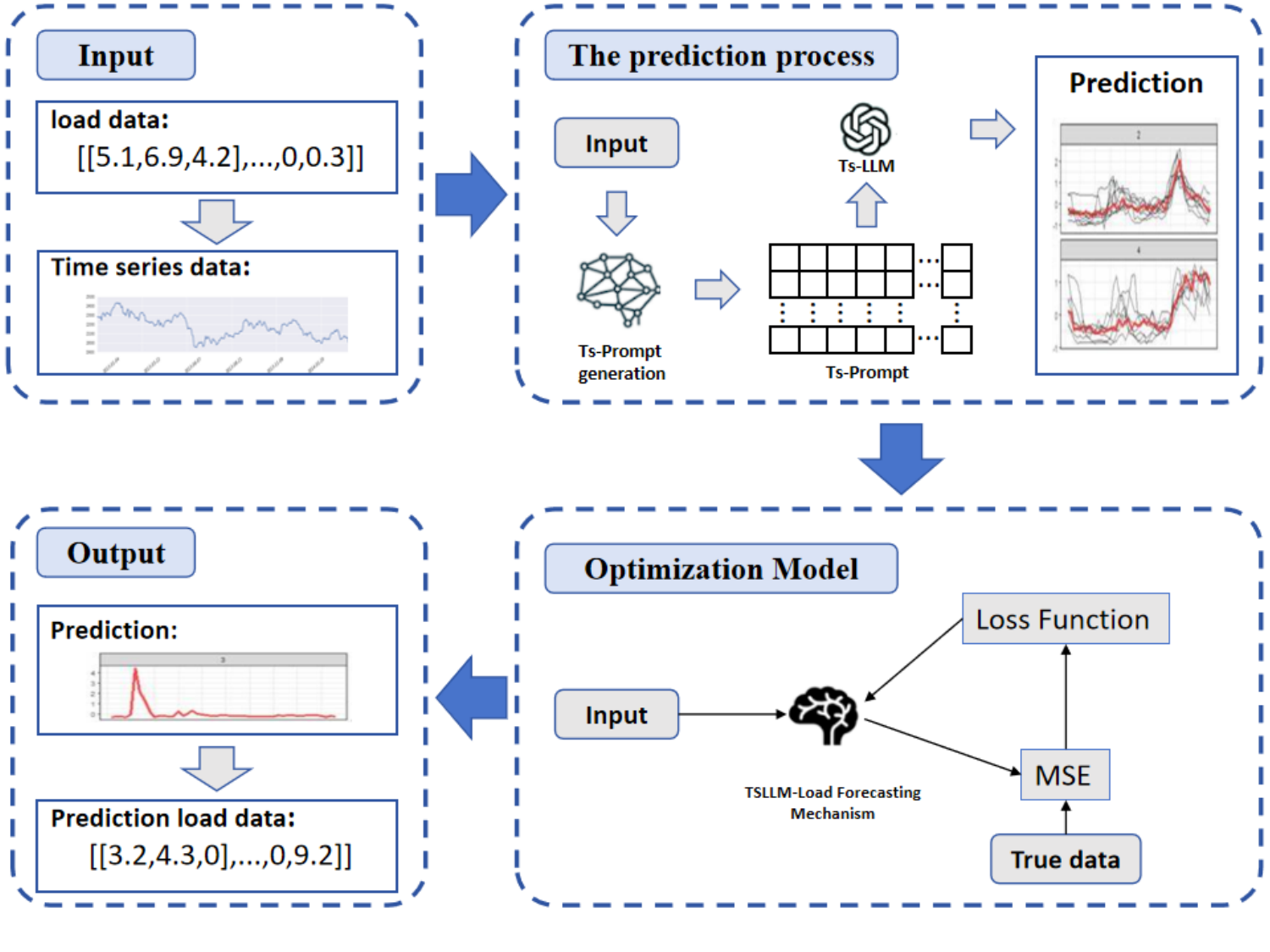}
\caption{Training process flowchart of the TSLLM-Load Forecasting Mechanism, illustrating parameter optimization and data flow through different components.}
\label{fig:training_flow}
\end{figure}

As illustrated in Fig.~\ref{fig:training_flow}, the training process adopts a selective optimization approach, focusing on specific model components while preserving the general capabilities of the pre-trained LLM. Empirical evidence suggests that maintaining most parameters in a non-trainable state yields superior generalization performance compared to comprehensive model retraining. Therefore, we selectively adjust the parameters of the position embedding layer and the layer normalization layer within the large language model, which are crucial for adapting the model to time series data while retaining its linguistic knowledge.

The optimization process targets five key components: the shared input layer in the time series prompt generation module, independent token extraction layers for each textual domain, position embedding layer, layer normalization layer, and the output layer of the large language model. The objective function is meticulously designed to achieve optimal prediction performance while ensuring effective alignment between time series embeddings and textual domain tokens:

\begin{equation}
\mathcal{L} = \mathcal{L}_{pred} + \lambda\mathcal{L}_{align}
\label{eq:loss_function}
\end{equation}

where $\mathcal{L}_{pred}$ represents the prediction loss implemented as Mean Squared Error (MSE) calculated as $\frac{1}{n}\sum_{i=1}^{n}(\hat{y}_i - y_i)^2$, and $\mathcal{L}_{align}$ denotes the similarity score matching function that aligns selected text tokens with decomposed and concatenated time series embeddings. The hyperparameter $\lambda$ serves as a balancing factor that regulates the contribution of the alignment term to the overall loss, with its value determined through systematic sensitivity analysis.

This strategically formulated training approach ensures that the model maintains its pre-trained language understanding capabilities while effectively adapting to the specific requirements of time series forecasting. The selective parameter optimization methodology promotes efficient learning while preserving the model's generalization ability, enabling robust performance in both conventional and zero-shot prediction scenarios.

\section{Results and Discussion}

\subsection{Experimental Setting}

To validate the effectiveness of the proposed algorithm, we conducted comprehensive experiments using the Australian solar household dataset~\cite{41}. This dataset comprises load data collected from 300 solar-equipped households within the Australian electricity grid jurisdiction from July 1, 2010, to June 30, 2013. The data encompasses three key measurements sampled at 30-minute intervals: solar panel generation load, major appliance load, and other appliance load.

Given experimental constraints, we randomly selected 20 households for zero-shot load forecasting evaluation using the TSLLM-Load Forecasting Mechanism. From these households, we designated one household's load data as the primary training target, partitioning its data into training, validation, and test sets with a ratio of 7:1:2. For the remaining 19 households, we extracted load data segments of equivalent length to the first household's test set as prediction targets. The implementation utilizes GPT-2 as the backbone pre-trained large language model. All experiments were conducted using PyTorch on a computing platform equipped with an NVIDIA GeForce GTX A100 80GB GPU.

\subsection{Experimental Framework}

\begin{figure}[!t]
\centering
\includegraphics[width=\linewidth]{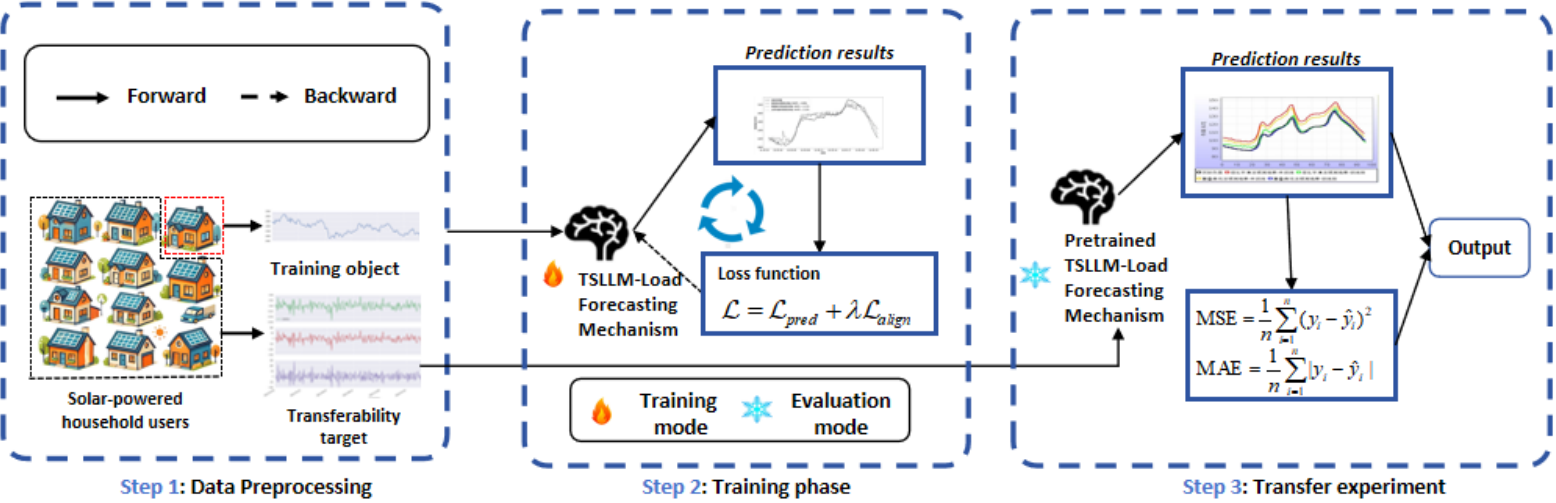}
\caption{Experimental workflow diagram showing the input processing, prediction generation, and output formation stages of the TSLLM-Load Forecasting Mechanism.}
\label{fig:exp_framework}
\end{figure}

The TSLLM-Load Forecasting Mechanism is designed to predict household electricity load for a 48-hour horizon. Fig.~\ref{fig:exp_framework} illustrates the experimental framework of the algorithm.

The model accepts three-dimensional input data covering the 10 days preceding the prediction time point: solar panel generation load, major appliance load, and other appliance load. These data streams are consolidated into a vector with dimensions $(512, 3)$ for model processing. The system transforms this input into a TS-Prompt comprising $n_{patch}$ data patch prompts, where each patch prompt consists of a prefix prompt $\mathbf{P}_{prefix}$ and patch embedding $\mathbf{P}_{embed}$.

\begin{figure}[!t]
\centering
\includegraphics[width=0.5\linewidth]{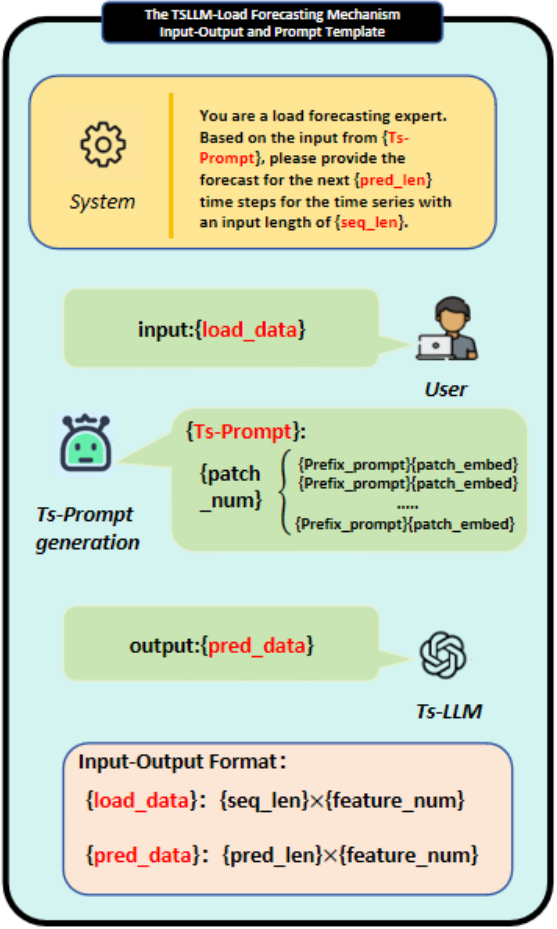}
\caption{Input-output structure and prompt template design of the TSLLM-Load Forecasting Mechanism, showing the transformation from raw data to prediction output.}
\label{fig:io_template}
\end{figure}

As shown in Fig.~\ref{fig:io_template}, the model generates load forecasts for the subsequent 48 hours without requiring training on historical data from the target household. The output maintains the three-dimensional structure, producing a $(96,3)$ vector representing the predicted loads across all three measurement categories.

To evaluate prediction performance, we employ two standard metrics: Mean Squared Error (MSE) and Mean Absolute Error (MAE). These metrics quantify the deviation between predicted values and actual measurements, defined as:

\begin{equation}
\text{MSE} = \frac{1}{n}\sum_{i=1}^{n}(y_i - \hat{y}_i)^2
\label{eq:mse}
\end{equation}

\begin{equation}
\text{MAE} = \frac{1}{n}\sum_{i=1}^{n}|y_i - \hat{y}_i|
\label{eq:mae}
\end{equation}

where $y_i$ represents the actual values, $\hat{y}_i$ denotes the predicted values, and $n$ indicates the number of samples. These metrics provide complementary perspectives on prediction accuracy, with MSE being more sensitive to large errors and MAE offering a direct interpretation of average prediction deviation.

\subsection{Comparative Analysis of Prediction Capability and Transferability}

To systematically evaluate the TSLLM-Load Forecasting Mechanism's effectiveness, we conducted a comprehensive comparative analysis against state-of-the-art baseline models. Our evaluation framework encompasses two key dimensions: prediction accuracy on the training dataset and transferability performance across multiple households.

First, we established a diverse baseline comprising advanced forecasting models: Informer \cite{42}, Autoformer \cite{43}, CNN\_LSTM \cite{44}, CNN\_LSTM\_Attention \cite{45}, and TCN\_LSTM\_Attention \cite{46}. Additionally, we included a single-task variant of our TSLLM-Load Forecasting mechanism to isolate the impact of the multi-task learning framework. Each baseline model represents a distinct approach to time series forecasting:

The CNN\_LSTM architecture combines convolutional feature extraction with sequential modeling through LSTM networks. The CNN\_LSTM\_Attention model enhances this architecture with an attention mechanism, enabling dynamic focus on significant sequence components. The TCN\_LSTM\_Attention model further incorporates temporal convolutional networks, leveraging causal and dilated convolutions for improved temporal pattern capture.

In the Transformer-based category, Informer introduces a probabilistic sparse attention mechanism optimized for time-series forecasting, while Autoformer extends this approach with adaptive frequency decomposition for separate modeling of seasonal and trend components.

\begin{figure}[!t]
\centering
\includegraphics[width=0.48\textwidth]{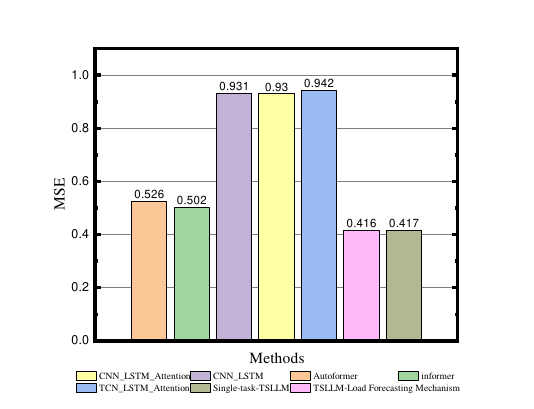}
\hfill
\includegraphics[width=0.48\textwidth]{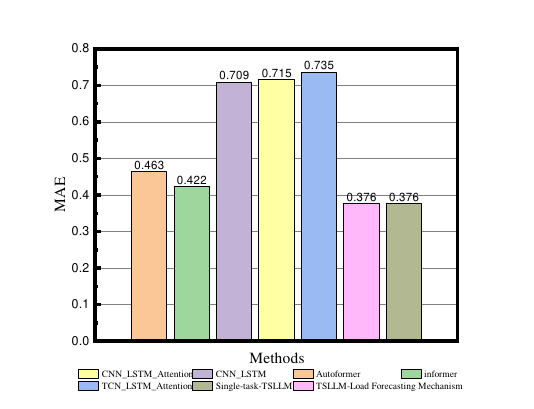}
\caption{Prediction results for training targets: (left) Mean Squared Error and (right) Mean Absolute Error comparisons across different models.}
\label{fig:prediction_results}
\end{figure}

\begin{figure}[!t]
\centering
\includegraphics[width=0.48\textwidth]{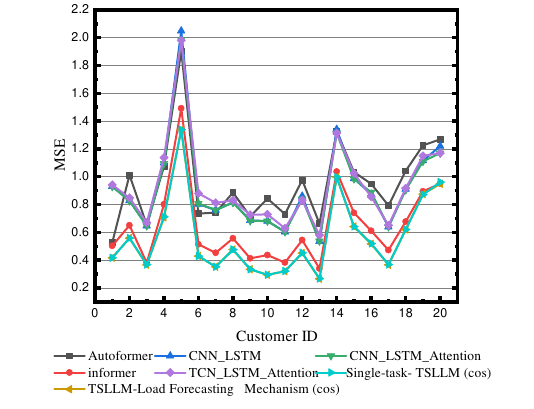}
\hfill
\includegraphics[width=0.48\textwidth]{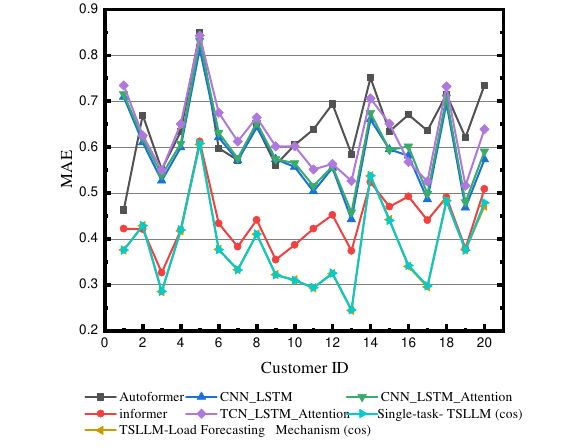}
\caption{Results of household power load transferability experiments: (left) Mean Squared Error and (right) Mean Absolute Error across all test households.}
\label{fig:transferability_results}
\end{figure}

Our experimental methodology ensured statistical robustness through three independent evaluation runs for each model, with results averaged across runs. The hyperparameter $\lambda$ was empirically optimized to 0.1 based on preliminary experiments. Detailed performance metrics are documented in Appendix A.

As illustrated in Fig.~\ref{fig:prediction_results}, our LLM-based approach demonstrates remarkable performance superiority on the training household dataset, significantly outperforming conventional forecasting methods. Moreover, in zero-shot prediction scenarios involving 19 household users, our method maintains consistent performance advantages. Notably, the MTL-based design consistently outperforms its single-task counterpart across most households, validating the effectiveness of our multi-task learning framework.

Compared to existing approaches, the proposed TSLLM-Load Forecasting Mechanism demonstrates substantial improvements in prediction accuracy across all baseline models. The comprehensive analysis reveals that our method achieves a 12.2\% reduction in total MSE compared to Informer, while showing even more significant improvements of 40.9\%, 38.7\%, 38.1\%, and 39.6\% when compared to Autoformer, CNN\_LSTM, CNN\_LSTM\_Attention, and TCN\_LSTM\_Attention, respectively. These consistent improvements across different baseline architectures underscore the robust performance advantages of our proposed approach. 

\subsection{Ablation Analysis}

To understand the contribution of individual components and validate our design choices, we conducted a systematic ablation study focusing on two critical aspects: the Multi-Task Learning (MTL) framework and the choice of similarity metrics. The ablation experiments maintained consistent conditions with $\lambda = 0.1$ while varying the learning framework (MTL vs. Single-task) and similarity metric (Cosine Similarity vs. Euclidean Distance).

\begin{figure}[!t]
\centering
\includegraphics[width=0.48\textwidth]{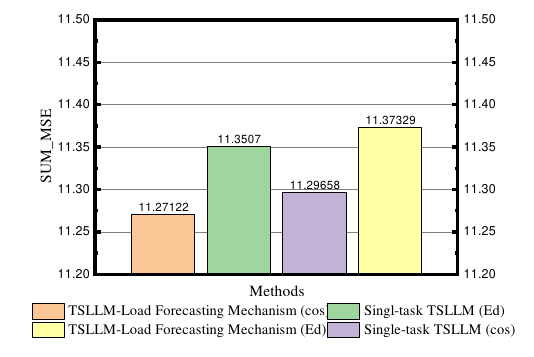}
\hfill
\includegraphics[width=0.48\textwidth]{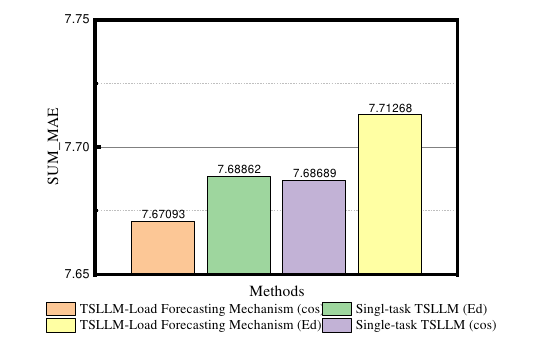}
\caption{Results of ablation experiments: (left) Mean Squared Error and (right) Mean Absolute Error comparisons between MTL and single-task variants with different similarity metrics.}
\label{fig:ablation_results}
\end{figure}

The ablation experiments reveal several important insights regarding the model's architectural design choices. The cosine similarity metric demonstrates consistently superior performance compared to Euclidean distance in both MTL and single-task implementations, with this advantage being particularly pronounced in the MTL framework. This observation suggests that cosine similarity more effectively captures the semantic relationships within the high-dimensional feature space. Furthermore, when employing cosine similarity, the MTL framework exhibits notably enhanced performance compared to its single-task counterpart, though this advantage becomes less pronounced when using Euclidean distance. This interaction between similarity metrics and learning frameworks indicates a beneficial synergistic relationship between MTL and cosine similarity. The optimal performance achieved through their combination demonstrates how these components effectively complement each other in capturing both feature relationships and task dependencies, validating our architectural design decisions.

These comprehensive findings provide strong empirical support for our architectural choices and demonstrate the importance of carefully considering both the learning framework and similarity metrics in multi-task time series forecasting applications. The results conclusively validate the effectiveness of our proposed approach in achieving robust and accurate load forecasting across diverse scenarios.

\subsection{Sensitivity Analysis}

To investigate the influence of hyperparameter $\lambda$ on model performance, we conducted a comprehensive sensitivity analysis using identical training and testing data while varying $\lambda$ values. This analysis evaluates how different weightings between prediction and alignment losses affect the model's forecasting accuracy.

\begin{table}[!t]
\caption{Sensitivity Analysis Results of Household Load Forecasting}
\label{tab:sensitivity}
\centering
\begin{tabular}{lccccc}
\toprule
Metric & $\lambda = 0.0$ & $\lambda = 0.01$ & $\lambda = 0.05$ & $\lambda = 0.1$ & $\lambda = 1.0$ \\
\midrule
Sum\_MSE & 11.8849 & 11.3626 & 11.3306 & \textbf{11.2712} & 11.8562 \\
Sum\_MAE & 8.0830 & 7.7422 & 7.7336 & \textbf{7.6710} & 7.8257 \\
\bottomrule
\end{tabular}
\end{table}

\begin{figure}[!t]
\centering
\includegraphics[width=0.7\textwidth]{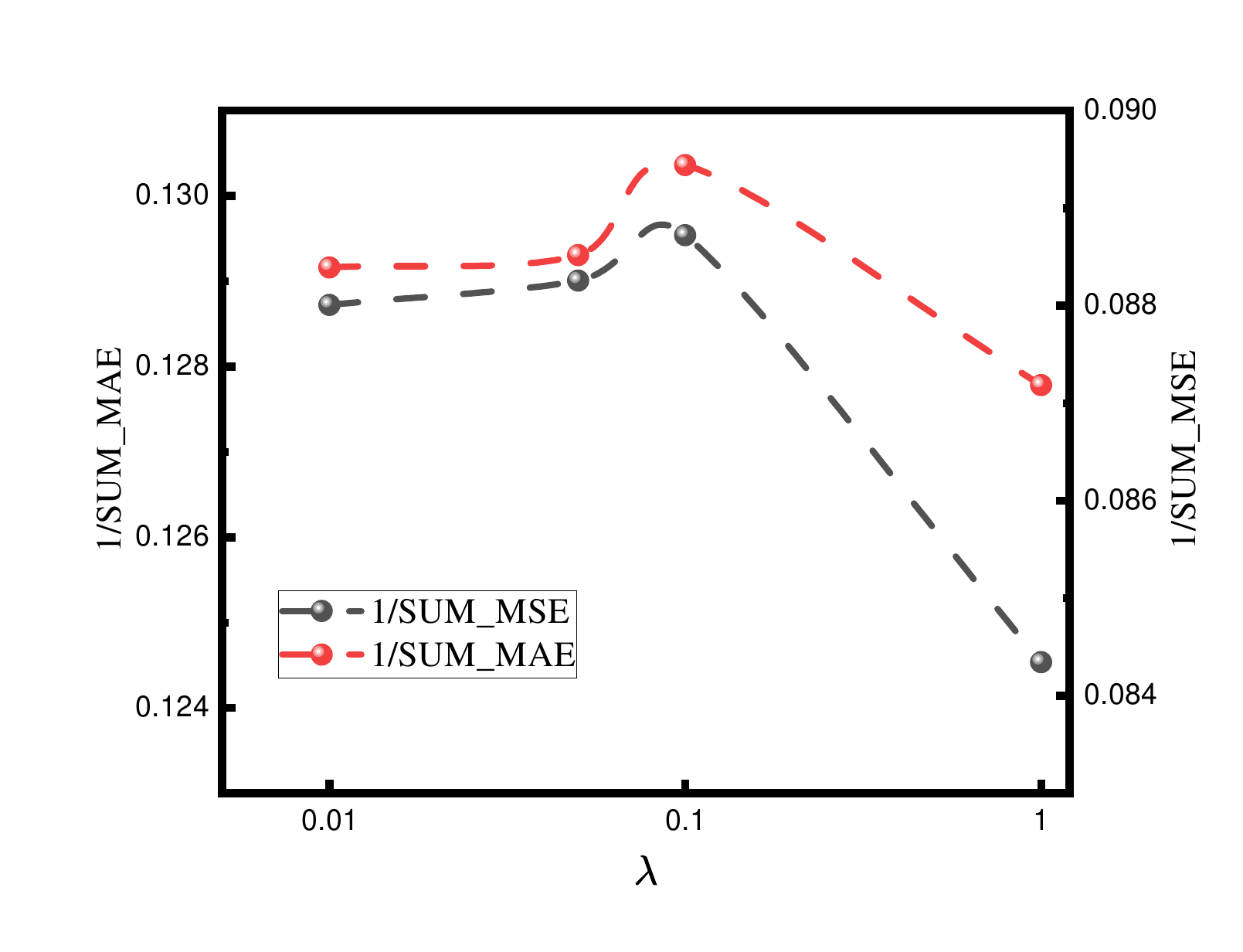}
\caption{Sensitivity analysis results showing the relationship between model accuracy (1/SUM\_MSE and 1/SUM\_MAE) and hyperparameter $\lambda$.}
\label{fig:sensitivity}
\end{figure}

The experimental results, as shown in Table~\ref{tab:sensitivity} and Fig.~\ref{fig:sensitivity}, demonstrate that increasing $\lambda$ initially improves time series prediction accuracy up to a certain threshold, beyond which prediction accuracy decreases significantly. The analysis reveals that while optimal $\lambda$ values differ slightly between MSE and MAE metrics, $\lambda = 0.1$ achieves consistently strong performance across both evaluation criteria. Figure~\ref{fig:sensitivity} illustrates this relationship, plotting the inverse of SUM\_MSE and SUM\_MAE against varying $\lambda$ values to provide a clear visualization of the model's sensitivity to this hyperparameter.

This sensitivity analysis provides crucial insights into the model's behavior and guides the selection of appropriate hyperparameter values for optimal performance. The results indicate that careful tuning of $\lambda$ is essential for balancing the contributions of prediction and alignment losses in the overall optimization objective.

\section{Conclusions}

Our paper introduces a method for load forecasting in power systems under zero-shot scenarios, termed the TSLLM-Load Forecasting Mechanism. This approach leverages large language models (LLMs) to address the limitations of existing forecasting methods, including poor transferability, inability to handle zero-shot scenarios without historical data, and lack of adaptability across diverse data sources.

To overcome the limitations of traditional methods in zero-shot forecasting scenarios, our approach combines large language models with load forecasting. To address the challenge of enabling LLMs to understand time series data, we designed a Time Series Prompt Generation Module. This module preprocesses power load data by decomposing and segmenting it into time series format. It employs a shared linear input layer and a task-independent text extraction layer to extract features from both the time series and text domains. Finally, a similarity alignment technique is applied to generate time series prompts, enabling the LLM to comprehend the time series data and perform load forecasting.

In a migration experiment involving load data from 20 Australian solar-powered households, the proposed method demonstrated superior performance compared to existing approaches in terms of Mean Squared Error (MSE) and Mean Absolute Error (MAE). It excelled both in testing on the training sample derived from one household's load data and in zero-shot migration experiments on the remaining 19 households' data. The method achieved a total MSE of 11.2712 and a total MAE of 7.6710 across the entire dataset, delivering at least a 12\% performance improvement over existing methods, thereby validating the accuracy and transferability of the proposed zero-shot forecasting method.

\section*{Acknowledgments}

This work is supported by the National Science Foundation of China under Grants 62203350 and 62373297, in part by Industrial Field Project---Key Industrial Innovation Chain (Group) of Shaanxi Province under Grant 2022ZDLGY06-02. The author gratefully acknowledges the support of K. C. Wong Education Foundation.

\appendix 





\bibliographystyle{elsarticle-num} 
\bibliography{reference.bib}

\end{document}